\documentclass{svjour3}                     % onecolumn (standard format)
\smartqed  % flush right qed marks, e.g. at end of proof
\usepackage[numbers]{natbib}
\usepackage{epsfig}
\usepackage{graphicx}
\usepackage{epsfig}
\usepackage{amsmath}
\usepackage{amssymb}
\usepackage{subfig}
\usepackage{epstopdf}
\usepackage{dsfont}
\usepackage{algorithm}
\usepackage{algorithmic}
\usepackage{booktabs}
\usepackage{multirow}
\usepackage{color}
\usepackage{siunitx}
\newcommand{\cbox}[1]{\raisebox{\depth}{\fcolorbox{black}{#1}{\null}}}
\usepackage{verbatim}
\usepackage{wrapfig}
\usepackage{hyperref}

\begin{document}

\title{Tracking using Numerous Anchor Points%\thanks{Grants or other notes
}


\author{Tanushri Chakravorty         \and
        Guillaume-Alexandre Bilodeau  \and 
      \'{E}ric Granger %etc.
}


\institute{Tanushri Chakravorty \and
            Guillaume-Alexandre Bilodeau \at
              LITIV Lab., Department of Computer and Software Engineering, Polytechnique Montreal, 				  Quebec, QC, Canada H3T1J4\\
              \email{tanushri.chakravorty@polymtl.ca}\\
              \email{gabilodeau@polymtl.ca}           %  \\
           \and
           \'{E}ric Granger \at
              LIVIA, \'{E}cole de technologie sup\'{e}rieure, Universit\'{e} du Qu\'{e}bec, Montreal, Quebec,
QC, Canada H3C1K3\\ \email{Eric.Granger@etsmtl.ca}
}

\date{Received: date / Accepted: date}

\maketitle
\begin{abstract}

In this paper, an online adaptive model-free tracker is proposed to track single objects in video sequences to deal with real-world tracking challenges like low-resolution, object deformation, occlusion and motion blur. The novelty lies in the construction of a strong appearance model that captures features from the initialized bounding box and then are assembled into anchor point features. These features memorize the global pattern of the object and have an internal star graph-like structure. These features are unique and flexible and helps tracking generic and deformable objects with no limitation on specific objects. In addition, the relevance of each feature is evaluated online using short-term consistency and long-term consistency. These parameters are adapted to retain consistent features that vote for the object location and that deal with outliers for long-term tracking scenarios. Additionally, voting in a Gaussian manner helps in tackling inherent noise of the tracking system and in accurate object localization. Furthermore, the proposed tracker uses pairwise distance measure to cope with scale variations and combines pixel-level binary features and global weighted color features for model update. Finally, experimental results on a visual tracking benchmark dataset are presented to demonstrate the effectiveness and competitiveness of the proposed tracker.
\end{abstract}

\section{Introduction}
\label{intro}
Visual object tracking can be considered as the task of detecting and locating an object of interest in a given video sequence. The object may undergo appearance variations due to illumination changes, occlusions, deformations, motion blur, etc. Also, the presence of similar looking objects (distractors) in the scene makes the tracking task more arduous. Despite the abundance of research on object tracking in the computer vision literature, there is no available full stack tracker that can address wide-range tracking challenges. Thus, there lies a scope of improvement for developing more accurate visual object trackers.

Domain specific applications like face tracking, human, pedestrian or hand tracking allows the algorithm designer to make some prior assumptions about the appearance of the object. Although well suited for some specific applications, they target specific objects. A tracker that can be generalized to a variety of objects is often more desirable. Therefore, building on such concepts lead to the notion of model-free trackers \cite{BMVC.20.6:abbreviated}. In such trackers, the initialization is performed in the first frame using a bounding box, and the sole information on the object to be tracked is derived from that first frame. Our proposed approach is a model-free tracker, where the initialization is performed using an axis-aligned bounding box.

In order to track an object efficiently, three aspects are crucial for any tracking process. First, building an appearance model that describes unique cues of the object such that it can be detected and tracked. Hence, the appearance model must consist of strong features that provide \textit{evidence} of an object's presence. Second, the appearance model should be \textit{flexible} for tackling appearance variations of the object. Finally, the appearance model should be updated at the \textit{correct} time so as to accommodate environmental changes due to illumination, scale, orientation etc. Therefore, a correct updating technique has to be determined to prevent erroneous features from being included in the appearance model.

In the proposed tracker, these three crucial aspects are considered, including a fourth crucial aspect related to the third i.e., preserving \textit{consistent} features in the appearance model for object localization, while removing \textit{inconsistent} features. In our tracker, the short-term and long-term consistencies of a feature are evaluated at every frame during the tracking process. Together, the long and short-term consistencies help to predict stable outputs and prevent the tracker from becoming overly sensitive to sudden changes in the environment. Moreover, including this fourth aspect also helps to track object in long-term tracking sequences, since the consistent features are retained in the model to locate the object. 

The rich representations and feature models provided by deep learning methods \cite{DBLP:journals/corr/WangLGY15}, \cite{NIPS2013_5192} are growing popular for visual object tracking and are delivering state-of-the-art results, however, they incur higher computational cost which is highly undesirable for tracking applications. On the other hand, simpler models based on color features and keypoints are capable of capturing distinct cues of the object, and perform equally well or sometimes even better than rich models in some scenarios \cite{7298823}. Therefore, thinking along those lines, we propose an \textit{anchor point} appearance model. Numerous keypoints on the object serve as anchor points, and are arranged in a structure defined with respect to the object center. Each keypoint predicts the object center location with its respective structure acting as anchor for the object center prediction. This structure of keypoints encoded with the object center helps to deal with occlusion and object deformation tracking challenges.

For deducing an accurate update strategy for a tracker, we believe that it is important to take advantage of both local and global features of the object. With the advent of binary feature descriptors like BRISK \cite{Leutenegger:2011:BBR:2355573.2356277} and FREAK \cite{Ortiz:2012:FFR:2354409.2354903}, it has become possible to find similar regions in an image at a lower computational cost. But, as they process larger image regions, it is difficult to identify local appearance changes at the pixel level. The LBSP (Local Binary Similarity Pattern) \cite{6836059} binary descriptor provides pixel level change detection. Instead of comparing patches, comparisons are done at the pixel level. To identify appearance changes at the global level, RGB color information is used. Taken together, binary descriptors and color information help in successfully updating of the appearance model because they prevent unwanted update at wrong time during the tracking process, for example during an occlusion, which might result in tracker drifts and track loss.

For accurate object localization, it is important to take account of inherent pixel noise of the tracking process. Particle filter-based methods like \cite{6836011}, \cite{DBLP:conf/iccv/MeiL09}, and motion-based methods like \cite{Shi:1993:GFT:866676}, are classic approaches for object localization, but do not consider the inherent pixel noise caused by local deformations during tracking. Hence, Gaussian prediction strategy proposed by \cite{ctse} can be utilized to deal with the above stated challenge as it helps to compensate for the keypoint feature displacement during scale change and fast motion of the object.

Finally, it is important to retain robust discriminant features in the appearance model for a tracker to be successful. Hence, an online method should be devised to determine the consistency of features during tracking. Thus, in our proposed tracker, for each feature, consistencies (long and short-term) are determined. The long-term consistency helps to retain consistent features for tracking and short-term consistency helps to control the sensitivity of the tracker to sudden appearance changes due to occlusion, illumination variation, etc.. Hence, consistent features should be kept in the model and others should be removed quickly or ignored temporarily.

The contributions of this paper can be summarized as follows. First, a new model-free tracker called TUNA (Tracking Using Numerous Anchor points) is proposed to track generic objects, with a novel appearance model that captures local and the global information about an object. This information is captured using numerous keypoint features that are assembled into \textit{anchor} points. They record the global structure of the object with respect to its center and the local information with its keypoint descriptor. Unlike other appearance models that emphasize on a single type of representation (either local or global), our model encapsulates both local and global features for a robust representation of an object. This new representation is distinctive and helps in dealing with distractors present in the environment. Second, a new updating strategy for appearance model is proposed using a combination of pixel level binary features and global level color features that determines the appropriate time for the anchor point appearance model update. Third, a novel technique is proposed for determining scale changes. Unlike other methods \cite{conf/iccv/HareST11}, \cite{Nebehay2014WACV} where transformation matrices are initially computed for adjusting scale, we propose a pairwise distance method between keypoint features for estimating scale change of the object. Finally, to preserve robust features for tracking, long and short-term consistencies of a feature are estimated and evaluated online during tracking. The long-term consistency aids in retaining consistent features for tracking and evolves (increase and decrease accordingly) with the tracking process, whereas the short-term consistency is evaluated instantaneously and aids in controlling the sensitivity of the tracker to sudden appearance changes due to occlusion, illumination variations, etc. Additionally, a strategy to deal with object deformation and occlusion is proposed with a Gaussian voting for accurate object localization.

The remaining of the paper is organized as follows. Section \ref{sec:RelW} describes the related research work in visual object tracking. Section \ref{sec:Idea} describes the concept of the proposed appearance model. Section \ref{sec:TUNA} and \ref{sec:TUNAadd} describe the tracking framework. Experimental results and analysis are presented in Section \ref{sec:eval}. Finally Section \ref{sec:con} draws the conclusions.

\section{Related Work}\label{sec:RelW}

In this section, different representations used by trackers to model the appearance of objects are presented. Generally, object representations can be classified into two broad categories, i.e. generative and discriminative. In generative representations, the object is modeled using features extracted from the object and then the object is matched by finding the most similar region compared to the model as in template matching trackers \cite{DBLP:conf/iccv/MeiL09}, \cite{1288530}. For example, Mean-Shift tracker \cite{854761} uses color features to find the object of interest, and Frag-Track \cite{Adam06robustfragments-based} models the object using histograms of local patches. Trackers like IVT \cite{IVT} use subspace models to incrementally learn the object representation. Sparse representation trackers like \cite{BaiyangLiu:2011:RTU:2191740.2191956}, \cite{CT}, consider a set of linear combination of templates to represent the object. The generative representations are usually less complex, but they are often unable to tackle the cluttered background scenes due to lack of background information included in the model, and might easily fail in such scenarios.

In contrast, discriminative representations consider tracking as a binary classification task. CSK \cite{Danelljan_2014_CVPR} uses color features and employs an online binary classifier for tracking. OAB \cite{BMVC.20.6:abbreviated} updates discriminative features via online boosting methods. Struck \cite{conf/iccv/HareST11} uses an SVM classifier to generate and learn the labels online for tracking and KCF \cite{henriques2015tracking} samples the region around the target. The cyclic shifts simulates translations of the target object. TLD \cite{TLD} uses two types of experts to train the detector online while tracking. The discriminative methods can tackle cluttered background scenes. However they are sensitive to noise because not a lot of information is available to train the classifier in the initial frame and therefore commonly suffer from tracking drift.

Part-based trackers \cite{SCMT}, \cite{ASLA} divide the object in smaller regions or patches, while \cite{ShuWang:2011:ST:2355573.2356430} uses superpixels as discriminative features and use learning to distinguish the object from background. The work of Cai et al. \cite{DBLP:conf/accv/2012-3}, proposes to decompose the object into superpixels and then use graph matching to find the association among frames. Due to their robust appearance representation using multiple parts, they provide useful cues during partial occlusion. On the other hand, they may not be able to handle object deformation due to the abrupt variation in translation of multiple parts.

Some trackers like CAT \cite{4538230} and SemiT \cite{SemiT}, use contextual information or supporting regions to deal with occlusion. But they might suffer from ambiguities due to the presence of several region of interest with their context. Some authors combine multiple features \cite{conf/eccv/YoonKY12}, or multiple trackers \cite{6126369}, to maintain multiple appearance models. An extensive summary on various appearance model representations and visual object trackers can be found in \cite{6671560} and \cite{Yilmaz:2006:OTS:1177352.1177355} respectively.

More related to our work are keypoint-based trackers SAT \cite{6836011}, CMT \cite{Nebehay2014WACV}, and CTSE \cite{ctse}. SAT \cite{6836011} uses a circular region for initializing tracking and computes a color histogram for that region. Further, keypoints are detected for the same region. For limiting the search region, it uses a particle filter framework for keypoint detection and matching for the next frame. It uses a histogram filtering method for estimating the quality of tracking. CMT \cite{Nebehay2014WACV} uses optical flow and consensus method that aid in finding reliable matches and hence improve tracking. CMT does not perform appearance update of the keypoint model. CTSE \cite{ctse} uses a structural configuration of keypoint features to track an object and refrain from updating the model. In contrast to previous keypoint-based tracking algorithms where the search region is limited, our proposed tracker searches the entire image for finding matches and verify these matches for mutual correspondence for higher reliability. This way our tracker can track object that have fast motion. The proposed appearance star graph-like model tackles object deformation due to appearance change of the object. Our method also introduces the concept of short and long-term consistencies of a keypoint feature. Together, the consistencies help to retain good features in the appearance model for object location and to predict stable outputs for object location by temporarily ignoring some keypoints present in the anchor point appearance model, yet keeping them in the model if they usually predict well.

\section{Ideation}\label{sec:Idea}

\begin{figure}[!htbp]
\begin{center}
%\fbox{\rule{0pt}{2in} \rule{0.8\linewidth}{0pt}}
 \includegraphics[width=\linewidth]{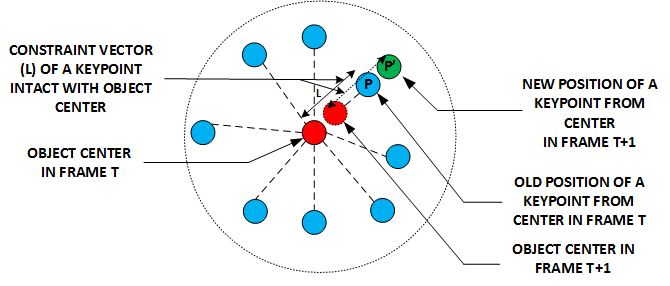}
\end{center}
   \caption{Anchor point Appearance Model. Note when the object moves, blue keypoint shifts to green (P to P'), its position changes but the encoded constrained vector $L$ is intact.}
\label{fig:am}
\label{fig:onecol}
\end{figure}

The model is inspired by deformable parts that has been used in the domain of object recognition and detection \cite{Felzenszwalb:2010:ODD:1850486.1850574}. In their method, the object is divided into smaller parts that are arranged in a star graph-like configuration. Each part is represented directly or indirectly in terms of other parts, and thus there is interaction among them. In our approach, the idea of interaction is slightly different. Here, the keypoints are described in relation to the center of the object by a vector (Refer Figure \ref{fig:am}). Thus, the keypoints are expressed in relation to the object center and not in terms of each other and thus can be considered as anchor points. Hence, except for object center position, no information is shared among the keypoints, which are unique and independent from each other. 

The interaction of keypoints with the center of the object is quite unique, as these keypoints belonging to the object bear a similar motion with respect to the object center. Our hypothesis is that keypoints with a constrained vector structure that have similar motion with respect to object center helps in predicting object's position in the next frame, because the encoded structure represents a strong feature of the object that has been already learnt with the help of anchor point features (Refer Figure \ref{fig:am}). Therefore, when the object moves in the next frame, the keypoints with respect to the object center will also move by the same spatial translation, keeping the constrained vector of these keypoints  approximately constant with the new position (P') of the re-matched keypoint as the reference. Hence, by re-detecting and matching the same keypoints for an object in the next frame, the new object position can be located. 

This model is robust to heavy occlusion as independent acting keypoints can be detected and tracked even if some keypoints become latent (not visible) during the tracking process. Our novel appearance model is efficient for tackling tracking challenges like distractors, occlusions (long and short), illumination variations, etc. because the keypoints with their structured vector point to the object center to locate it and vote with their short-term and long-term consistencies. The long-term consistency is adapted online for a keypoint feature and aids in retaining good learnt keypoint features in the anchor point appearance model, whereas the short-term consistency is an evaluation of a prediction response by a keypoint feature for current frame. Therefore, even if some keypoints become latent, still the location can be predicted using other visible keypoints. The short-term and long-term consistencies associated with a keypoint act as a feature learning memory. The voting by an anchor point for the object center is performed using a gaussian window, which compensates for the keypoint displacement during object deformation. Further, the constrained vector is distinctive and tackles with distractors and background. Finally, the proposed model is not limited to specific objects and thus can be applied to a wide range of embedded vision robotics and surveillance applications.

\begin{figure}[!htbp]
\begin{center}
%\fbox{\rule{0pt}{2in} \rule{.9\linewidth}{0pt}}
 \includegraphics[width=\linewidth]{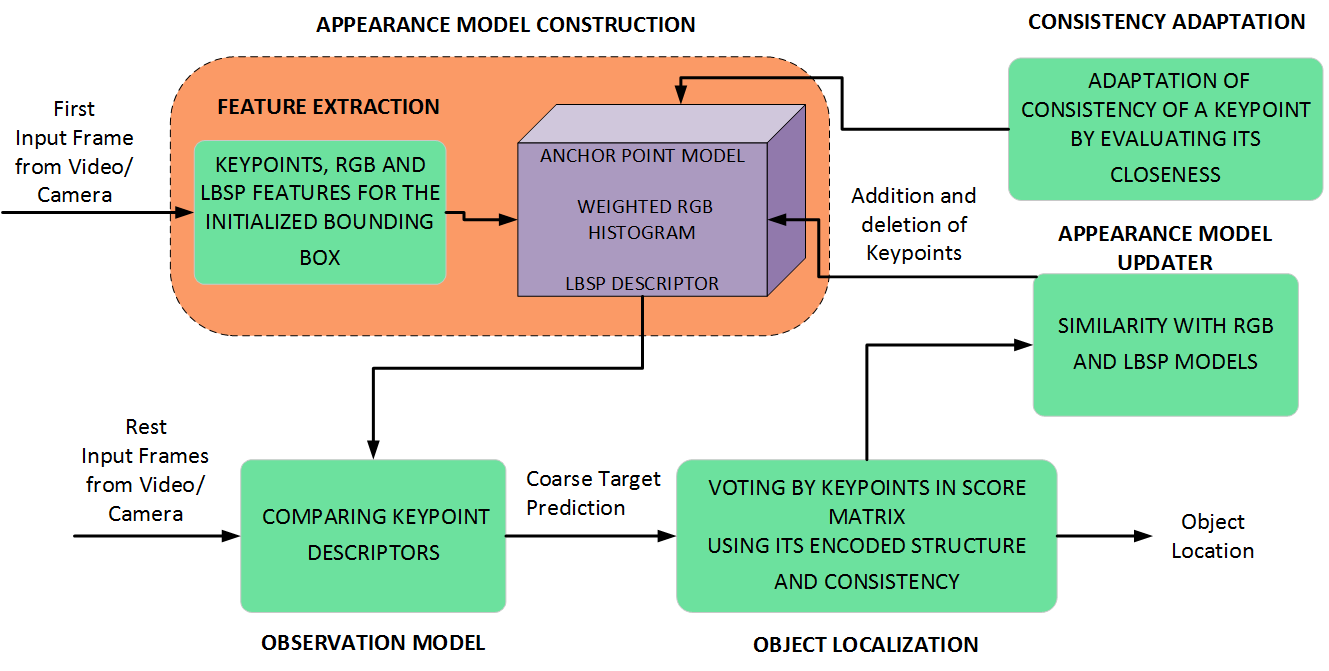}
\end{center}
    \caption{Tracking using Numerous Anchor points (TUNA)}
\label{fig:blockDia}
\end{figure}

\section{Tracking Using Numerous Anchor points}\label{sec:TUNA}

In this section our proposed tracker called TUNA (Tracking Using Numerous Anchor points) is detailed. Figure \ref{fig:blockDia} represents the block diagram of our tracking system.
The main system components are \textit{feature extractor}, \textit{appearance model}, \textit{observation model}, \textit{object localization}, \textit{consistency adaptation} and finally \textit{appearance model updater}.

The term \textit{anchor point} refers to a keypoint and vector pointing to the object center, along with its consistencies. The tracking is executed as follows. In the first frame keypoints are extracted and described for the initialized bounding box. These keypoints are modeled in a star-like structure with the object center as the root of the tree and their vectors (Euclidean distances in X and Y) with respect to the center are encoded (Refer Figure \ref{fig:am}). With this step, the construction of the \textit{anchor point appearance model} is completed. At the same time, the global model is built by computing pixel-level LBSP and color RGB reference models. Then, keypoint features are detected and described in the next frame and are matched for similarity with the keypoints present in the anchor point appearance model. This is the \textit{observation model}, where the keypoint descriptors are matched for similarity using $L2$ norm.  Then, each matched keypoint votes with its associated anchor point and its present location for the object center for the current frame. For \textit{object localization}, all the individual votes are analyzed for maximum aggregation of votes, which represent the final object position. The \textit{consistency adaptation} reflects the consistency of prediction of anchor points present in the model. The \textit{long-term} consistency evolves over the tracking process and becomes larger if a keypoint is re-matched and predicts closer to the final target center and vice-versa. While the \textit{short-term} consistency prevents abrupt change of object location predictions due to dynamic appearance changes. The \textit{appearance model updater} computes for maximum similarity between the final tracking output obtained in previous step with the RGB and LBSP appearance models for deciding if the model should be updated or not. In this step, new anchor points (keypoint features with their vector and consistency) are added to the anchor point model and poor keypoint features are removed from it based on their consistencies.

\subsection{Feature Extraction}
In this step, \textit{three} features are extracted viz., anchor point features (SIFT \cite{sift} keypoints encoded with a vector pointing to the center of the object), color (RGB) and pixel level binary features (LBSP \cite{6836011}). First, keypoints are detected and described for the bounding box and encoded into anchor points. SIFT keypoints are used as they are proven robust to illumination, rotation, scale etc. \cite{sift}. Any other keypoints can be used. Similarly, RGB histogram and LBSP descriptors are computed for the object contained in the bounding box. The LBSP is a 16-bit binary coded descriptor and provides pixel level modeling. For the RGB color model, a weighted 3-D histogram for all the pixel values lying in the initialized bounding box is calculated. Hence, in the proposed tracking framework, three features are kept as reference models for the object to be tracked. For object localization only anchor point features are used whereas the color and pixel-level features are used during the anchor point appearance model update.

\subsection{Anchor Point Appearance Model}

The filtered keypoints obtained from the initialized axis-aligned bounding box, can be visualized in the form of a directed star graph-like structure denoted as $G(P,L)$, where vertices are directed towards the center. $P$ represents the keypoints belonging to the object (Refer Figure \ref{fig:am}) and edge $L$, represents the connection between the vertices and the root of the structure. In our scenario, the vector of a keypoint is an edge, and is denoted as $L=[\Delta x_{k_{i}}]$, directed towards the center. Here, $\Delta x_{k_{i}}$ contains the Euclidean distance of the keypoint's location $x_{k_{i}}$ with respect to the center. Hence, the anchor point appearance model consists of the following:\\

 \begin{itemize} 
 \item Descriptor of keypoint in the anchor point model\\
\vspace{-0.3cm}
 \item Constraint vector of a keypoint that describes its location with respect to the object center Consistencies denoted by $L$\\
\vspace{-0.3cm}
 \item ST (short-term) and LT (long-term) consistencies of a keypoint that indicates the keypoint's relevance for the object. A keypoint located nearby to the object center will have higher LT consistency as compared to others and is adapted online with a learning parameter during the tracking process. Further, the keypoint's ST will have a higher value if its individual prediction for the object center is closer to the globally voted object localization by all the keypoints present in anchor point model.
\end{itemize}

\subsection{Observation Model}
After the construction of the appearance model in the first frame, the keypoints are detected and described for the subsequent frame. Detecting keypoints all over the frame helps in finding an object having large or abrupt motion. Then, these keypoints are matched for similarity with the feature descriptors of the keypoints present in the anchor point appearance model by comparing their feature descriptors using $L2$ norm. For filtering bad keypoint matches, the ratio test of \cite{sift} is used, and the matches that have a distance ratio of more than 0.9 are removed. Moreover, the mutual matching correspondence of keypoints between consecutive frames is confirmed, i.e, one-sided matched keypoints are not considered for voting and object localization. Only two-sided mutual matches are kept.

For the rest of the text in the paper, the matching of a keypoint will refer to matching of keypoint in the current frame with those keypoints present in the anchor point appearance model.

\subsection{Object Localization}
\begin{figure}[!htbp]
\begin{center}
%\fbox{\rule{0pt}{2in} \rule{0.8\linewidth}{0pt}}
 \includegraphics[width=0.9\linewidth]{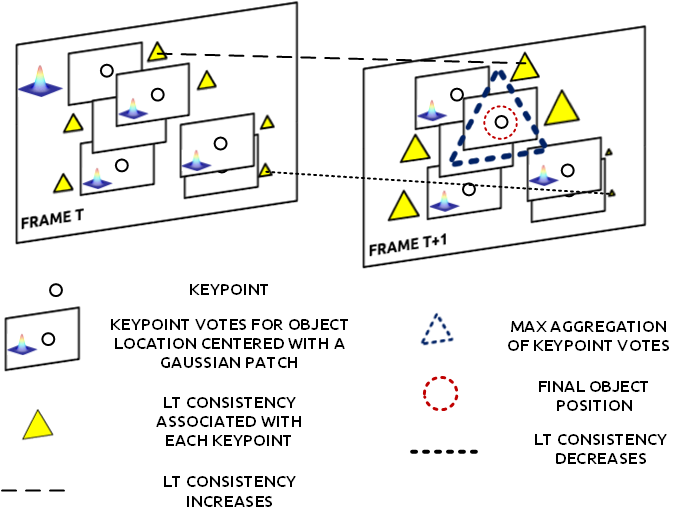}
\end{center}
   \caption{Visualization of keypoint voting and object localization. Here yellow triangles represent the consistency of a keypoint. Bigger yellow triangles represent higher consistency and vice-versa.}
\label{fig:voting}
\label{fig:onecol}
\end{figure}	
Consider visualizing the voting by a keypoint for the object center in the image space (Refer Figure \ref{fig:voting}). The pixel location at which the keypoint is pointing for the object center, is centering a Gaussian patch which gives more value to the center than other pixel locations around it. The advantage of voting in a Gaussian patch is that it allows to localize the object center even if the keypoint gets displaced from its original configuration in the anchor point appearance model and thus, is flexible towards deformation of object. Therefore, when a keypoint $k_{i}$ is matched, it votes for the object center $x$, with its structured constrained vector (${L_{k_{i}}}$) as : %Equation \ref{eq:pdf}.

\begin{equation}\label{eq:pdf}
P(x | k_{i}) \propto \frac{1}{\sqrt{2 \pi |{\Sigma}|}}exp(-0.5(x - ({L_{k_{i}}} + x_{k_{i}}))^{T}{  \Sigma }^{-1 }(x - ({L_{k_{i}}} + x_{k_{i}}))
\end{equation}

Here, $P(x | k_{i})$ is the constraint vector score given by a keypoint, $k_{i}$ for the object center, $x$, and $\Sigma$ is covariance.  

\begin{figure}[!htbp]
\begin{center}
%\fbox{\rule{0pt}{2in} \rule{0.8\linewidth}{0pt}}
  \includegraphics[width=1.1\linewidth]{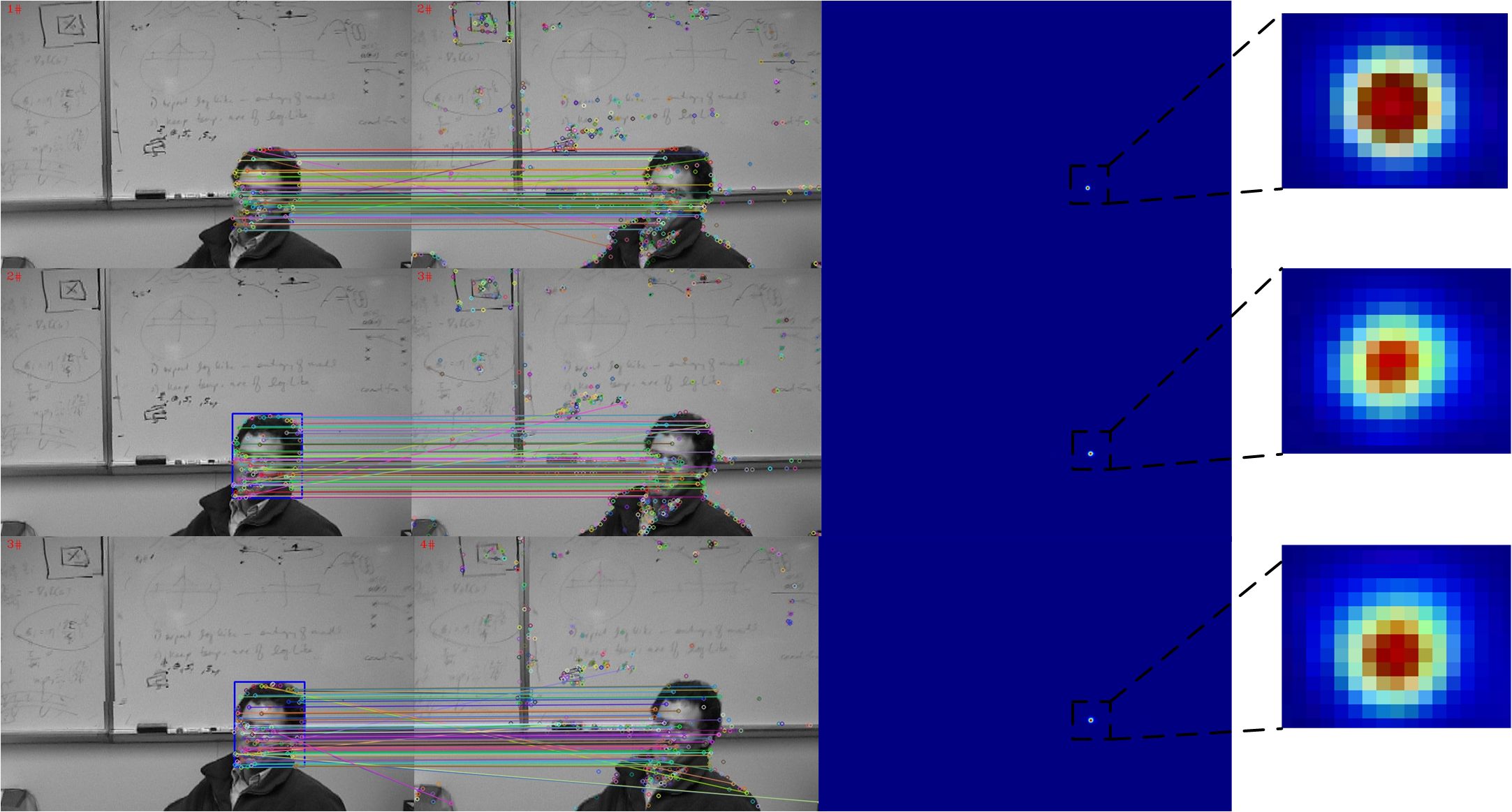}
\end{center}
   \caption{Illustration of keypoint matching between consecutive frames and their corresponding Score Matrix. The red color signifies more votes for the object center. Please note that for the pixels that are farther from the center, the values decrease gradually.}
\label{fig:votingMag}
\end{figure}

Hence, each keypoint votes for the object center with its constrained vector score, its long-term consistency, and its short-term consistency as a total score in a Score Matrix, $SM$. Therefore, the total score for the object center can be formulated as a likelihood function, which is given by the dot product of the constrained vector score of a keypoint, its long-term consistency, and its short-term consistency. Hence, the likelihood expression as a function of total score by a keypoint can be written as :

\begin{equation}\label{eq:mScore}
SM(x)= \sum_{i=0}^{K}P(x | k_{i}).LT_{C_{k_{i}}}.ST_{C_{k_{i}}}I_{(k_{i}\in K)}
\end{equation}

where, $LT_{C_{k_{i}}}$ is the long-term consistency, $ST_{C_{k_{i}}}$ is short-term consistency of a keypoint, and $I_{(k^{(i)}\in K)}$ is an indicator function, which is set for keypoints contained in the anchor point appearance model that are matched in current frame. $K$ is the total number of keypoints present in the anchor point appearance model. The cluster where the sum of individual scores is highest is taken as the final object center location, denoted as $x_{OCenter}$. The cluster shown a dashed blue colored triangle in Figure \ref{fig:voting} represents that majority of keypoints are voting for the same object location. For better understanding, Figure \ref{fig:votingMag} illustrates the matching of keypoints between consecutive frames and their voting in the Score Matrix for object localization. The dark red represents the predicted object center and has the highest value in the Score Matrix. Hence, the final object location is given by:

\begin{equation}\label{eq:fScore}
x_{OCenter}=\operatorname{arg\,max}\left ( SM(x) | x \in SM\right )
\end{equation}

\subsection{Model Parameter Estimation}\label{subsec:PEst}

\paragraph{Long-term (LT) consistency of a keypoint}\label{par:lt}: It is estimated using a measure called \textit{closeness}, $M_{C_{k_{i}}}$ associated with a keypoint, and is measured by computing the proximity of a keypoint's prediction for object center denoted as $x_{{PredCenter}_{k_{i}}}$, with respect to the final obtained object center using Equation \ref{eq:fScore}. It is calculated using Equation \ref{eq:long-term} for the current frame $T$.

\begin{equation}\label{eq:long-term}
M_{C_{k_{i}}}^t  = max((1 - |\alpha (x_{OCenter} - x_{{PredCenter}_{k_{i}}})|),0.0)
\end{equation}

Hence, a keypoint that predicts closer to the center will have higher closeness value, as compared to others. The keypoints which predicted very far from the final obtained center are assigned a value of $0.0$, thus reducing their impact on voting for the object center for the future frames. This parameter is adapted for all the frames as we will see in the next subsection.
For the initial frame, closeness measure for all the keypoints present in the appearance model are initialized using Equation \ref{eq:closenessInit} as:

\begin{equation}\label{eq:closenessInit}
M_{C_{k_{i}}}^{t_{0}} = max((1 - |\alpha*L_{k_{i}}^{0}|),0.5)
\end{equation}

Here, $L_{k_{i}}^{0}$ is the initial vector associated with each keypoint ${k_{i}}$ for frame $T_{0}$, and $\alpha$ is closeness factor. In the first frame LT consistency of a keypoint equals to $M_{C_{k_{i}}}^{t_{0}}$. The motive of using such an initialization function is to help in assigning larger closeness value to those located keypoints that lie closer to the object center (indicating that the keypoints probably belong to the object) as compared to those which are farther (indicating that they may belong to the background). Thus, $0.5$ is assigned to pixels that are farther from the center instead of zero so that they can be still considered for object location. Moreover, since it is the first frame, there lies is a slight degree of uncertainty.

\paragraph{Short-term (ST) consistency of a keypoint}\label{par:st}:
By analyzing how far away the keypoint predicted from the final object center obtained in frame $t$, the impact of ${k_{i}}$ for future object center predictions in voting can be controlled. For instance, if a keypoint's prediction for the object center is very close to the object center obtained from Equation \ref{eq:fScore} in frame $t$, then its short-term consistency for frame $t+1$ increases using Equation \ref{eq:st}. But on the other hand, if a keypoint voted far from the object center then its short-term consistency for prediction for object center reduces for frame $t+1$. The advantage of analyzing short-term consistency is that it aids in coping with sudden appearance changes of the object due to occlusion, rotation, illumination etc. For instance, if a keypoint has a high long-term consistency and due to sudden appearance change, the keypoint votes incorrectly for the object center with a higher voting score, its short-term consistency will be lower, therefore reducing its impact globally for the voting score in Equation \ref{eq:fScore} :

\begin{equation}\label{eq:st}
ST_{C_{k_{i}}}^{t+1} =  exp\left ( -\frac{(x_{{PredCenter}_{k_{i}}}^{t} - x_{OCenter}^{t})^2}{\eta} \right )
\end{equation}

where $\eta$ is a scaling factor.

\subsection{Model Parameter Adaptation}\label{subsec:PAdap}

In this step, the long-term consistency of a keypoint is adapted for all the keypoints that are present in the appearance model depending on their closeness measure. Keypoints that are matched more often, and for which their individual prediction is closer to the majority prediction obtained from Equation \ref{eq:fScore}, will have larger closeness as compared to the rest of the keypoints that are predicting farther. This also provides an indication that whether the keypoint belongs to the object or the background, since if a keypoint does not predict for the center or if it is predicting very far, its \textit{closeness} will be less and its long-term consistency will reduce eventually, according to Equation \ref{eq:closeAdapt}.

\begin{equation}\label{eq:closeAdapt}
LT_{C_{k_{i}}}^{t+1}= 
\begin{cases}
    (1 - \delta)LT_{C_{k_{i}}}^{t}  + \delta M_{C_{k_{i}}}^t,& \text{if } I_{(k_{i}\in K)} \text{is true} \\
  	(1 - \delta)LT_{C_{k_{i}}}^{t},              & \text{otherwise}
\end{cases}
\end{equation}   

where, $\delta$ is an adaptation factor.

\begin{figure}[!htbp]
\begin{center}
%\fbox{\rule{0pt}{2in} \rule{0.8\linewidth}{0pt}}
 \includegraphics[width=0.8\linewidth]{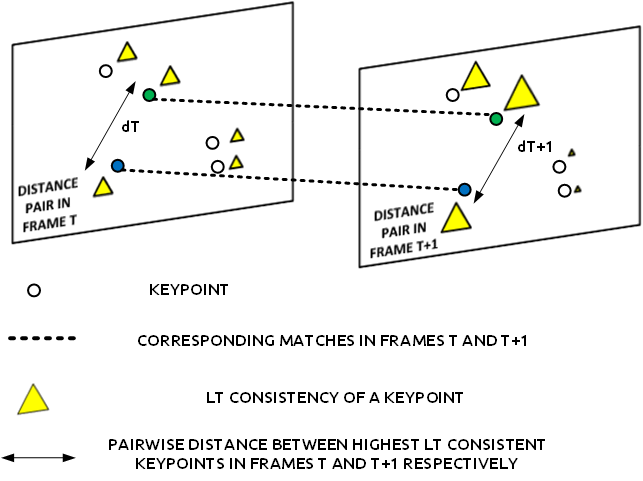}
\end{center}
   \caption{Scale Estimation}
\label{fig:scale}
\end{figure}

\subsection{Appearance Model Update}
Finally, the appearance model is updated only when a high tracking quality is achieved. The criteria for measuring the tracking quality is based on two features, viz., the local pixel level LBSP (Local Binary Similarity Pattern) feature, and the global RGB color feature respectively. Only the anchor point appearance model is used for object localization and it is \textit{updated} during tracking process based on matching similarity criteria of LBSP and RGB color features that are kept as \textit{reference} models from the initial frame. Hence, after every object location by the tracker using the anchor point model, the LBSP and RGB color models for the obtained bounding box are matched for similarity with their respective LBSP and weighted RGB reference models. The LBSP descriptor is matched for similarity using \textit{Hamming} distance and the weighted color histogram is matched for similarity using $L2$ norm respectively. The advantage of having a weighted color histogram, is to give more importance to the foreground pixels that are closer to the object center and less importance to the background pixels. 

If the similarity comparisons agree with the reference models, then new anchor points are added to the anchor point appearance model. The newly added keypoints are initialized with their respective structured constrained vectors and consistency values. The keypoints whose long-term consistency is poor and is lower than a threshold of $LT_{C_{min}}$ are removed from the model. 

\subsection{Scale Estimation}
To adapt the scale to the current object location, we utilize a pairwise distance measure between keypoints that have been matched for similarity between two consecutive frames. This Euclidean paired distance represents the distance between keypoints and indicates how much the keypoint has moved due to the scale change of the object. Moreover, by taking a mean of these paired distances, a single computed scale value can be applied to the bounding box. The number of keypoints that are considered for computing the pairwise distance depends on the total number of matches between two consecutive frames and their long-term consistency. The distance between the keypoint having the highest consistency (represented by blue color in Figure \ref{fig:scale}) with all other keypoints (represented by green color) are computed for frame $T$. Similarly, their corresponding distance is noted in frame $T+1$. 

Then, a distance ratio is computed for a keypoint pair and is given by $d(T+1)/dT$ and a mean value is computed. The final scale change is applied to the bounding box after a period of every ten frames. Moreover, it is only applied when the mean lies within $\pm$ 10 \% of the initial size of the target object. This is because, we assume that the scale of the object would not undergo such an abrupt difference in scale between two consecutive frames. Note, the scale estimation is not limited to a fixed aspect ratio of the object.

\section{Additional details on the working of TUNA}\label{sec:TUNAadd}

Due to partial occlusion, some keypoints become latent (not visible) during the tracking process. Therefore, only the keypoints having indicator function, $I_{(k^{(i)}\in K)}$, as one can be tracked. These keypoints act independently for object prediction and vote for the object center with their vector and their consistencies. Together the LT and ST consistencies associated with features helps in voting for object localization, since the consistent performing features only vote in the score matrix with their associated consistencies. 

For some frames, if there are no matches due to a long-term occlusion, motion blur or an out-of-plane rotation, the last obtained object location is not updated until the object appears again and the consistent keypoints present in the anchor point model start predicting again. Refraining from updating the location during this time helps in making less location errors. Together, the LT and ST consistencies prevent abrupt prediction changes when the object undergoes large appearance variations during tracking. For example, it may be possible that a background keypoint having LT consistency is present in the anchor point model, and is predicting wrongly for the object center. But, while evaluating its ST consistency, its value is lower for the next frame, since it predicted farther from the object center obtained using Equation \ref{eq:mScore}. Hence, when it votes again for object center in the next frame with its consistencies, the voting score reduces in the score matrix for the next frame. This is because the LT consistency reduces due to its adaptation by learning factor, according to Equation \ref{eq:closeAdapt} and the ST consistency reduces, according to Equation \ref{eq:st} respectively. This helps in preventing erroneous object location predictions.

Further, during object deformation some keypoints may get displaced, therefore when a keypoint votes for the object centering a Gaussian patch, the gaussian acts a flexible window for the keypoint displacement. Hence, the higher value assignment to the center as compared to rest of the pixels surrounding the keypoint makes TUNA tolerant to deformations. Further, the anchor point appearance model with constrained vector is distinctive and helps to deal with distractors and background because the model captures the pattern of the local information of the object using keypoint descriptor and the global information of the object with the keypoint constrained vector.

\section{Evaluation}\label{sec:eval}
The tracker performance is evaluated on  a recent benchmark \cite{Wu_2013_CVPR} having 51 video sequences. The video sequences have several attributes like severe illumination changes, abrupt motion changes, object deformations and appearance changes, scale variations, camera motion, long-term scenarios and occlusions. Our results are compared against other classic tracking algorithms: Multiple Instance Learning (MIL) \cite{MIL}, Color-based Probabilistic tracking (CPF) \cite{CPF}, Circulant Structure of tracking-by-detection with Kernels (CSK) \cite{CSK}, Kernel-based object tracking (KMS) \cite{KMS}, Semi-supervised on-line boosting for robust Tracking (SemiT) \cite{SemiT}, real-time Compressive Tracking (CT) \cite{CT}, Beyond Semi-Supervised Tracking (BSBT) \cite{BSBT}, Robust Fragments-based Tracking using the integral histogram (Frag) \cite{Adam06robustfragments-based}, Tracking-Learning-Detection (TLD) \cite{TLD}, Mean-Shift blob tracking through Scale space (SMS) \cite{Mean-Shift},  Online Robust Image Alignment via iterative convex optimization (ORIA) \cite{ORIA}, visual tracking via Adaptive Structural Local sparse Appearance model (ASLA) \cite{ASLA}, and Incremental learning for robust Visual Tracking (IVT) \cite{IVT}, respectively.

\subsection{Quantitative Evaluation}
\begin{figure}[!htbp]
\centering
\subfloat
{\includegraphics[width=0.7\columnwidth]{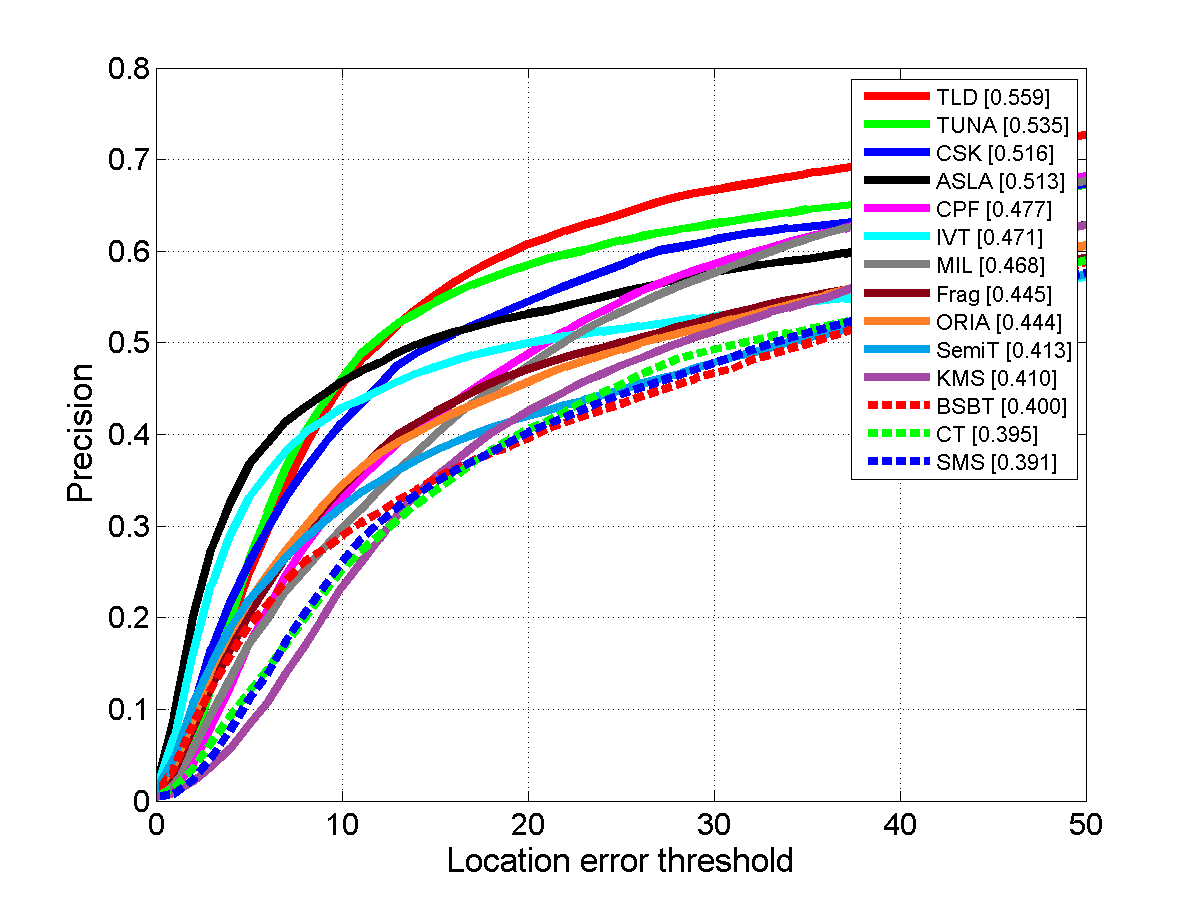}}\\\vspace*{-1.5em}
\subfloat
{\includegraphics[width=0.7\columnwidth]{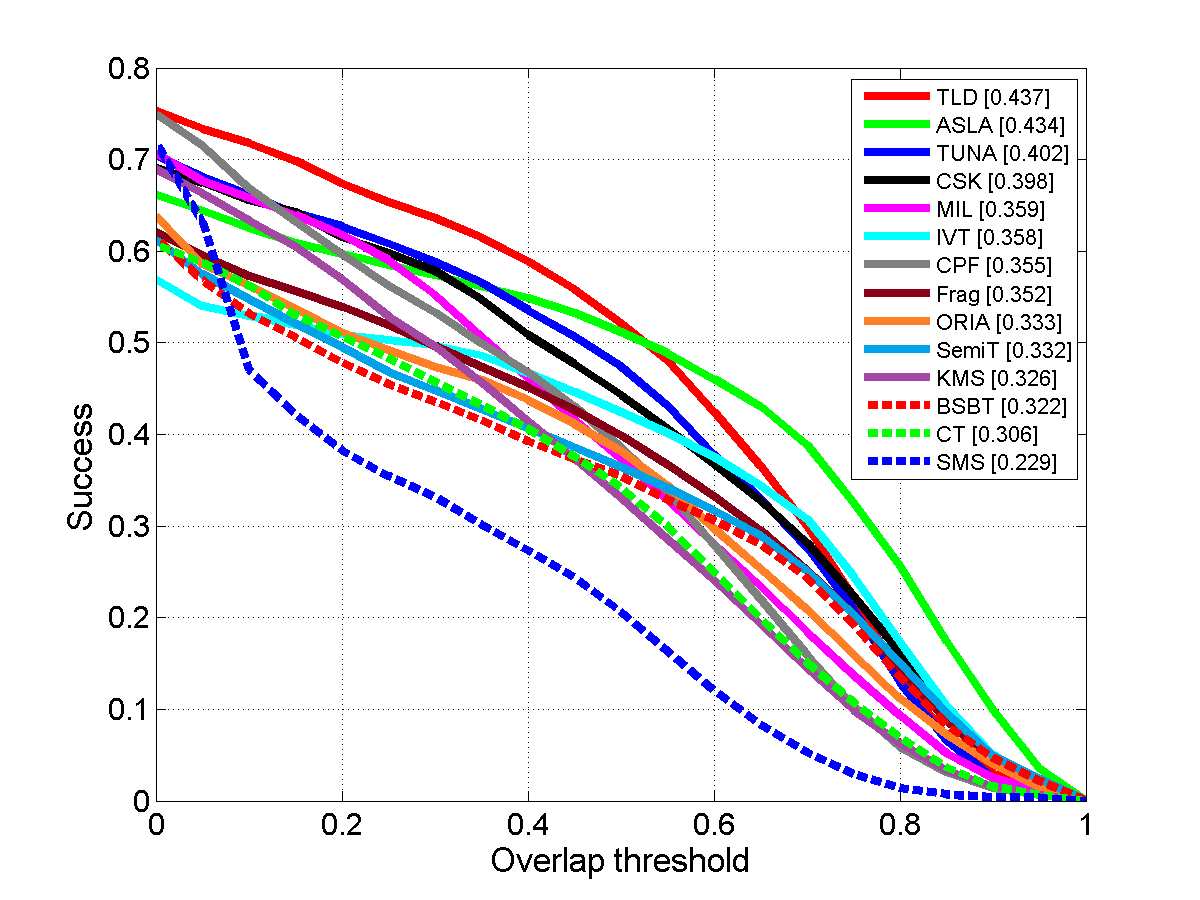}}
\caption{Precision and Success plots on all 51 video sequences. The proposed tracker TUNA outperforms several other state-of-the-art trackers. Best viewed in color and zoomed in.}
\label{fig:7}
\end{figure}

The evaluation is done using the standard evaluation protocol suggested by \cite{Wu_2013_CVPR}, which uses two criteria. The first is precision, where position error between the center of the tracking result and that of the ground truth is used. A threshold of 20 pixels is used for ranking the trackers. This threshold represents the percentage of frames for which the tracker was less than 20 pixels from the ground truth. 
The second is success that represents the bounding box overlap of the tracking result with the ground-truth.  The overlap is the ratio of intersection and union of predicted bounding box with the ground-truth bounding box. Instead of using the standard threshold of 0.5, this benchmark uses AUC (Area Under Curve) and the threshold is varied from 0 to 1 and the AUC across all the thresholds is reported as success results. A larger AUC indicates higher accuracy of the tracker.

\begin{table}[h]
\centering

\begin{tabular}{p{0.48\linewidth} p{0.2\linewidth} p{0.1\linewidth}}

\toprule

{\textit{Algorithm}}  & {\textit{Overall Precision}} & {\textit{AUC}} \\

\midrule

{\textbf{TUNA (Proposed)}} \\

Anchor Point Model + LBSP & 53.0\% & \textbf{40.9}\% \\
Anchor Point Model + RGB & 51.7\% & 38.1\% \\

 Anchor Point Model + LBSP + RGB & \textbf{53.5}\% & 40.2\% \\
{\textbf{TUNA (Without Scale)}}  & 52.4\% & 39.9\% \\

\textbf{CSK} \cite{CSK} & 51.6\% & 39.8\% \\

\textbf{MIL} \cite{MIL} & 46.8\% & 35.9\% \\

\textbf{TLD} \cite{TLD} & \textit{\textbf{55.9}}\% & \textit{\textbf{43.7}}\% \\

\textbf{Frag} \cite{Adam06robustfragments-based} & 44.5\% & 35.2\% \\
\bottomrule
\end{tabular}
\caption{Summary of Experimental Results on the 51 video dataset. The bold italic represents the best results and bold represents the second best results.}\hspace*{-5.0em}
\label{tab:sum}
\end{table}

We tested three versions of our proposed tracker TUNA viz., first using anchor point model for object localization and LBSP features as reference model for appearance update, and the second using anchor point model for object localization and RGB feature as reference model for appearance and third one using anchor point model for object localization and LBSP and RGB features, both as reference models for appearance update. We remark that by using the third version, the overall precision of the tracker increases (Refer Table \ref{tab:sum} and Figure \ref{fig:7}). In addition, when tested without scale estimation, the precision and success of the tracker reduces a little. TUNA performs second after TLD which emphasizes that detection module is an important engineering component in the tracker. TUNA\footnote{https://bitbucket.org/tanushri/tuna} is implemented in C++ using OpenCV 3.0.0 library\footnote{http://opencv.org/}. It runs at a mean FPS of 8 (computed over the  51 sequences) on Intel Core i7 @ 3.40 GHz, 8GB RAM computer. The parameters used in all experiments for TUNA are summarized in Table \ref{tab:summaryP}. The approximate computational complexity of TUNA is quadratic. The complexity can be attributed to the matching of keypoints between two frames as $O(k_{1}k_{2})$, where $k_{1}$ represents the number of keypoints in the appearance model and $k_{2}$ represents the number of keypoints detected in the next consecutive frame. Additionally, voting by keypoints for center location and finding the maximum in the SM (Score Matrix) is $O(n^2)$, where $n$ is the size of the image.

\begin{table}[htbp]
\centering
\caption{Parameters used in all Experiments}
\label{tab:summaryP}
\begin{tabular}{p{0.45\linewidth} p{0.3\linewidth}}

\toprule

{TUNA Parameters}  &{Value}  \\

\midrule

Closeness Parameter & $\alpha$ = 0.005 \\

ST Consistency Parameter &  $\eta$ = 5000.0 \\

LT Consistency  Initialization & $\lambda$ = 0.5 \\

LT Consistency Adaptation & $\delta$ = 0.1\\

LT Consistency Min. Threshold &  $LC_{min}$ = 0.1\\
\bottomrule
\end{tabular}
\end{table}

\vspace{-1mm}

\subsection{Attribute Wise Analysis}

As seen from Table \ref{tab:CLE}, low resolution (LR) severely affects the performance of most of the trackers. But our proposed tracker TUNA performs the best among all, showing the superiority of the anchor point model. Even in videos with low resolution, keypoint features can be extracted and thus encoding the structure of the object. Hence, keypoints with the structure votes accurately for the object location. Unlike other trackers that perform poorly, CSK that uses densely sampled features, can cope up.

Among all the other attributes, TLD performs better than other trackers showing the importance on its re-detection and failure module engineered in the tracker. Nevertheless, such component can further improve the performance of our proposed tracker, but TUNA still proves its superiority over TLD in low resolution (LR) and performs competitively on other challenging sequences performing as second best.

TUNA is able to perform very well in video sequences having motion blur (MB) due to the following facts. As each keypoint votes for the object location is associated with LT and ST consistencies, which are adapted during the tracking process, it helps to avoid too many wrong predictions. For instance, if a keypoint is LT consistent but if its ST consistency is too low, its voting contribution in score matrix for object location reduces. This also indicates, that a keypoint from background (or an outlier) might be predicting for the object center wrongly, if it is included in the model. Thus, it is better to have few good predictions rather than having too many false predictions for object location. Moreover, maintaining a holistic color model and local pixel level helps in preventing unwanted model update, therefore preventing the model from drifts. TUNA also performs well on videos having fast motion (FM) as keypoints are detected all over the frame. Therefore, matching for object location is performed on a larger search region, unlike ASLA where the search region is limited due to the particle filter. 

For videos with occlusion (OCC), TLD performs best due to its re-detection scheme. Note that the color features prove their distinctiveness for occlusion with CPF tracker. TUNA performs competitively here ranking as third among others. This is because even if some keypoints become hidden due to occlusion, the independent acting keypoints in the anchor point model votes for the object center with their consistencies. Moreover, the keypoints from the background will have smaller LT consistency and smaller voting contribution as compared to the foreground keypoints. Hence, there are fewer chances for incorrect object prediction during partial occlusion.

The proposed tracker is able to handle object deformation (DEF) very well. This is because when a keypoint is matched, it votes with the anchor points (that has the constrained vector structure of a keypoint) centered with Gaussian patch. Hence, even if the keypoints get displaced due to object deformation, the gaussian patch allows voting in a neighborhood with more emphasis on the center pixel, which makes it handle the error associated with the keypoint deformation. For some frames, if there are no matches due to long-term occlusion and out-of-plane rotation, the obtained object location is not updated until the object appears again and the keypoints start predicting, thus making erroneous location errors.

The pairwise ratio distance between keypoints helps to gauge the scale change between two frames accurately by analyzing the LT consistencies of keypoints. Moreover, the scaling technique does not take into account any fixed aspect ratio and thus can be applied to objects of various sizes. TUNA ranks third among the state-of-the-art trackers for scale variations (SV). The videos with background clutter (BC) also impacts the performance of all trackers except CSK, showing dense sampling of negative features around the object helps to better discriminate the object from background.

\subsection{Tracking behavior during the entire duration of the video sequences}

\begin{figure}[!htbp]
\captionsetup[subfigure]{labelformat=empty}
\centering
\subfloat[(a)]{\includegraphics[width=0.5\linewidth]{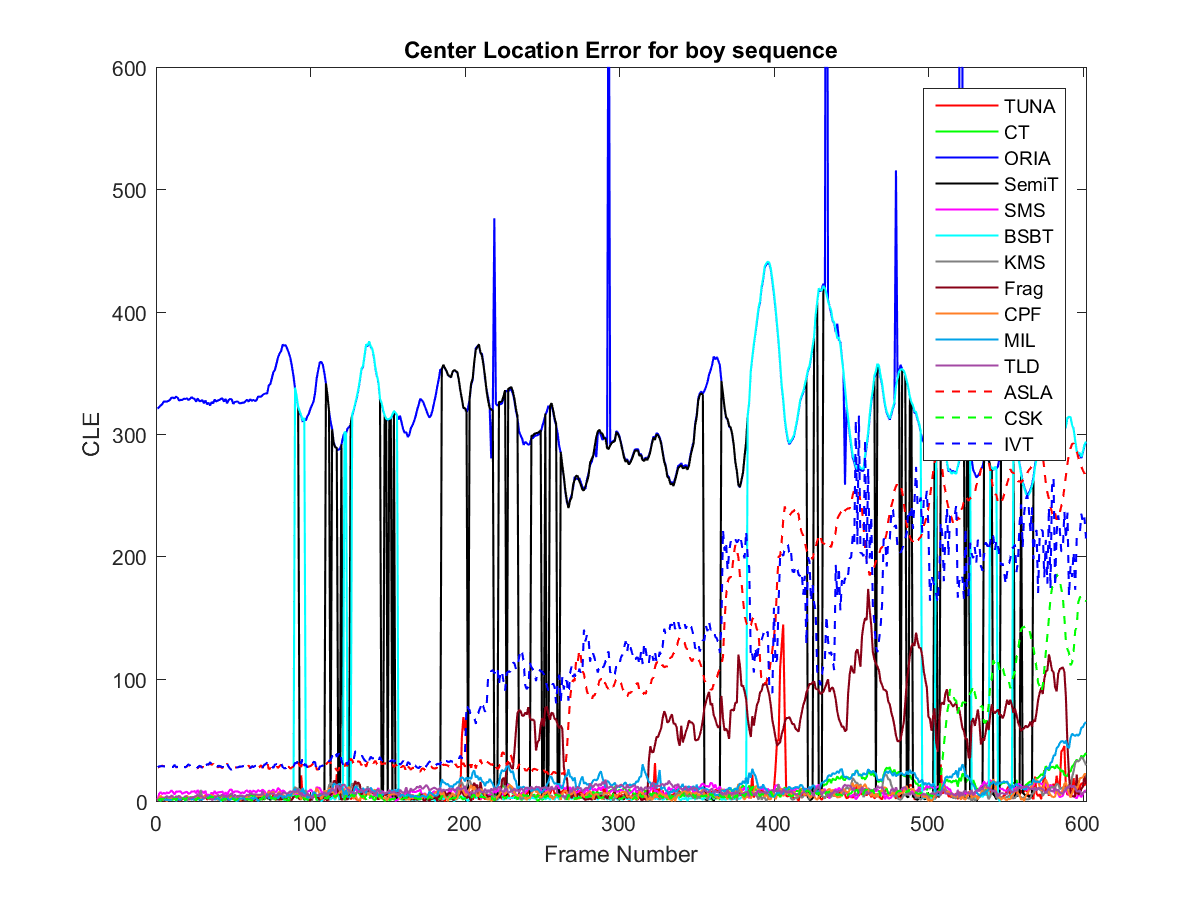}}
\subfloat[(b)]{\includegraphics[width=0.5\linewidth]{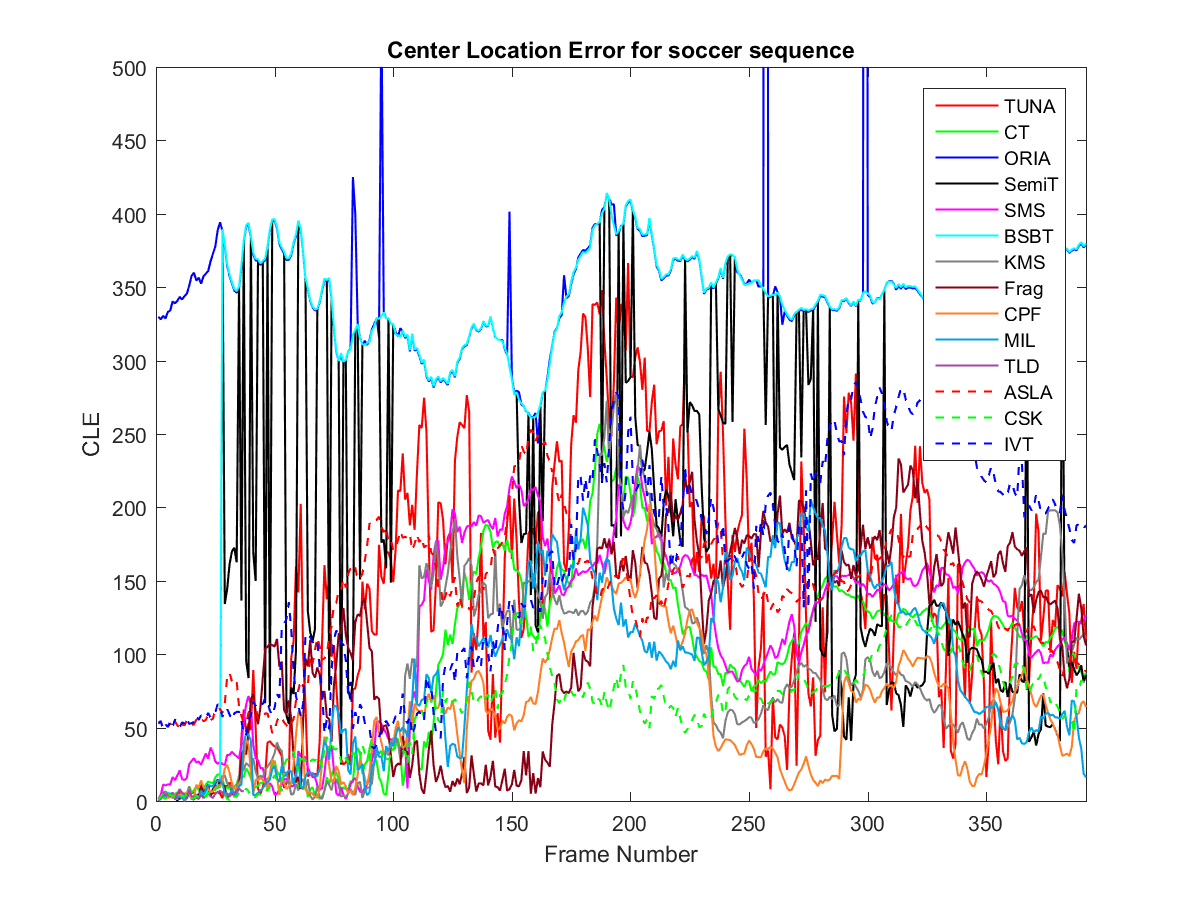}}\\
\subfloat[(c)]{\includegraphics[width=0.5\linewidth]{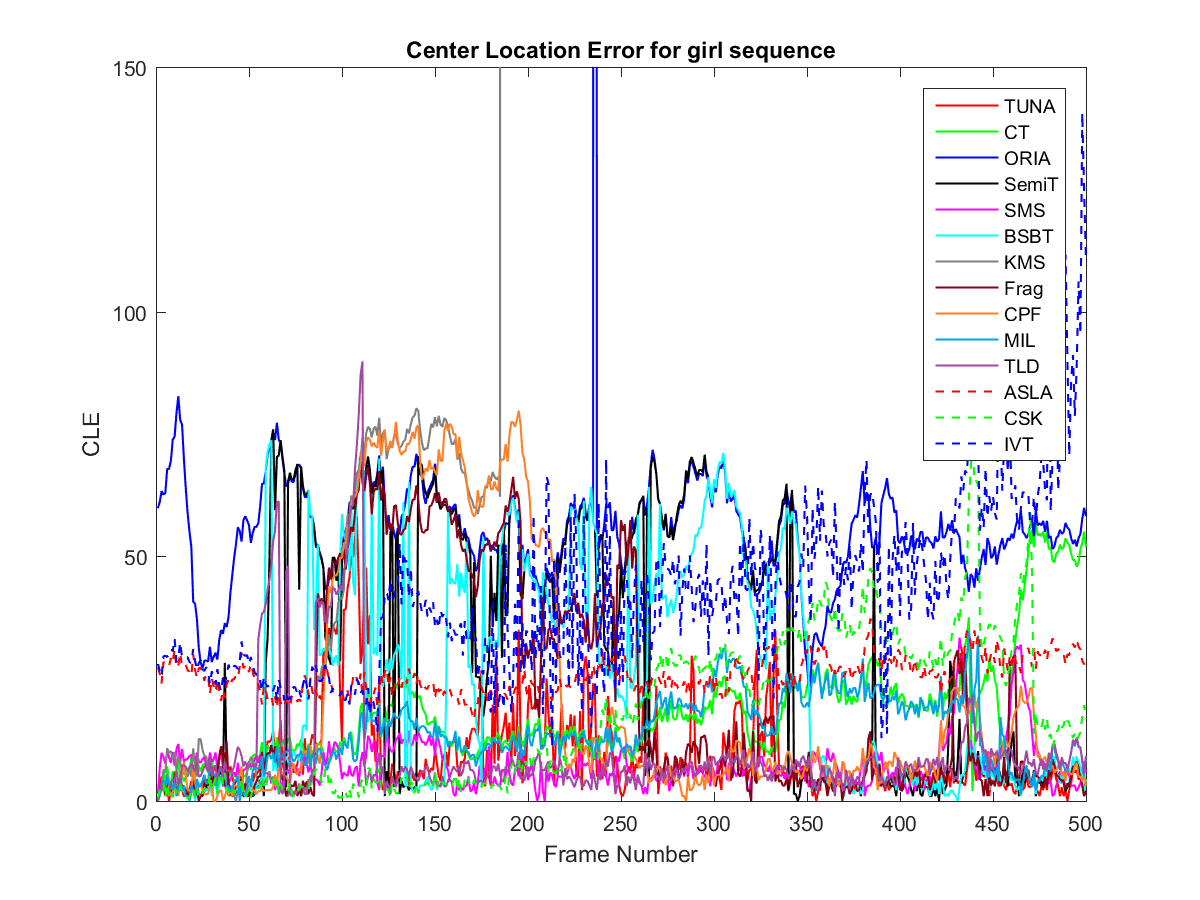}}
\subfloat[(d)]{\includegraphics[width=0.5\linewidth]{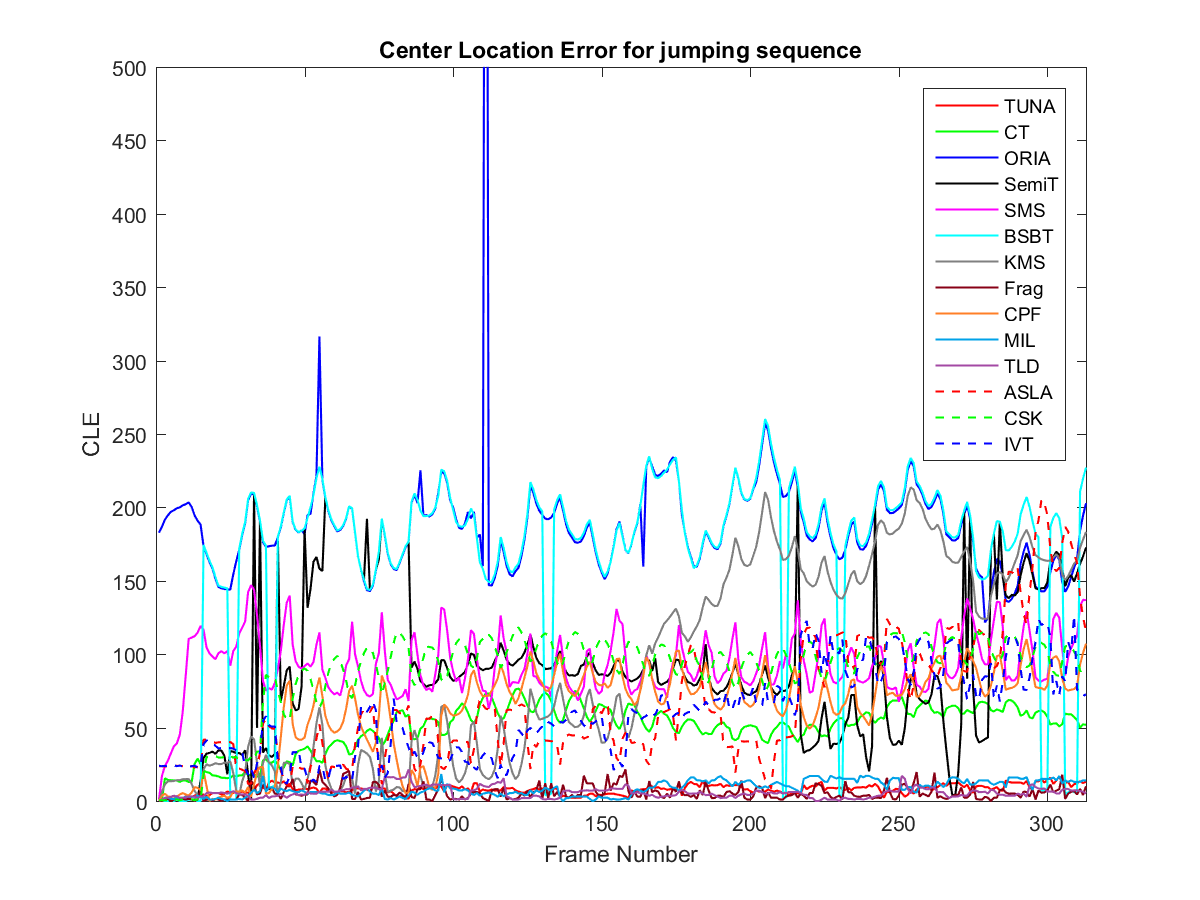}}\\
\caption{\footnotesize{Tracking behavior of the trackers for the entire video sequences: (a) boy (b) soccer (c) girl (d) jumping (best viewed when zoomed in) }}
\label{fig:behav}
\end{figure}

Figure \ref{fig:behav} shows the tracking results for the Center Location Error (CLE) on the entire duration of the video sequences: boy, soccer, girl and jumping respectively. These video sequences have several tracking challenges like in-plane rotation, out-of-plane rotation, fast motion of the object, highly cluttered background etc..The CLE is computed as the Euclidean distance between the center location of the tracker's output (bounding box) with the center location of the bounding box of the ground truth. It can be seen that TUNA is able to track during the whole duration of sequence and does not loose track of the object, as compared to others which might have drifted or lost track during challenges like large object motion, both in-plane and out-of-plane rotation, background clutter except the soccer sequence. It is a sequence which contains a highly cluttered background and almost all the trackers have a large variation from the ground-truth center.

\subsection{Qualitative Evaluation}

\begin{figure}[!htbp]
\subfloat
{\includegraphics[width=0.5\columnwidth]{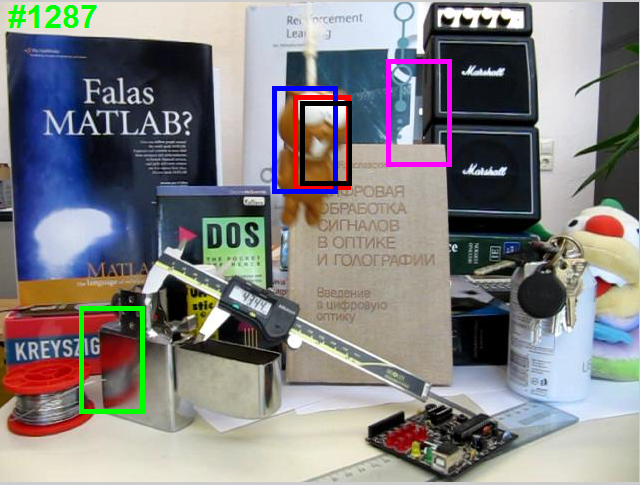}}
\subfloat
{\includegraphics[width=0.51\columnwidth]{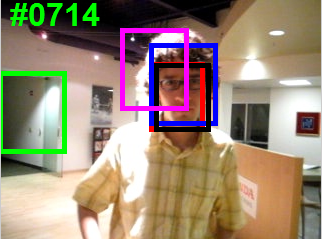}}\\
\\
\subfloat{\includegraphics[width=0.5\columnwidth]{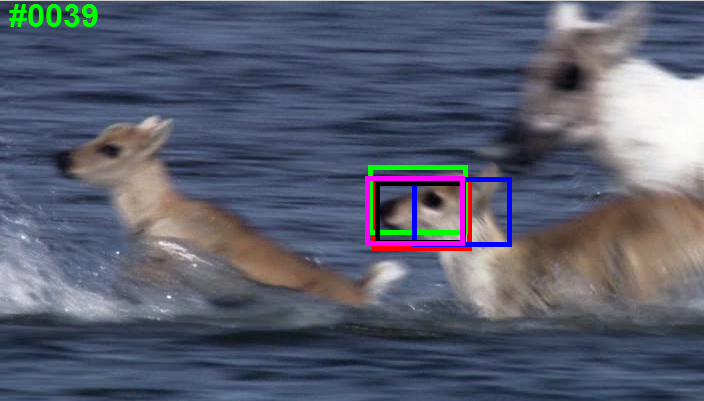}}
\subfloat{\includegraphics[width=0.505\columnwidth]{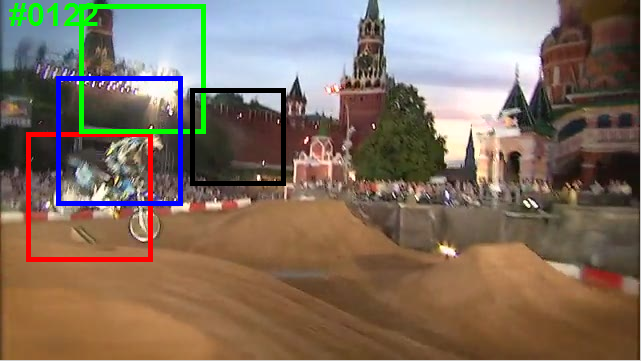}}
\\
\subfloat
{\includegraphics[width=0.48\columnwidth]{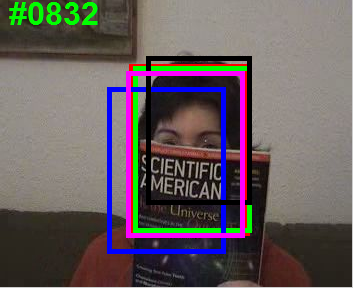}}
\subfloat{\includegraphics[width=0.527\columnwidth]{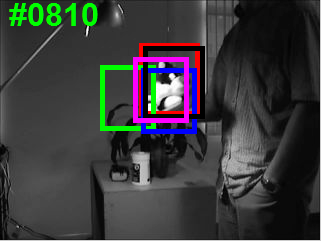}}\\
\subfloat{\includegraphics[width=0.5\columnwidth]{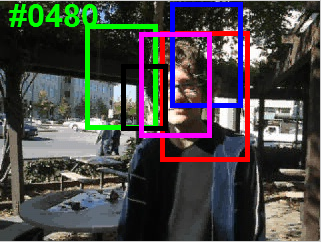}}
\subfloat{\includegraphics[width=0.5\columnwidth]{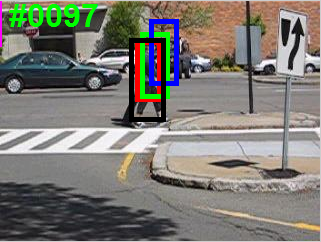}}
\caption{Snapshots results of selected tracking algorithms on video sequences: \textit{lemming}, \textit{david}, \textit{deer}, \textit{motorcycling}, \textit{faceocc1}, \textit{sylvester}, \textit{trellis} and \textit{couple} respectively.}
\subfloat{
   \hspace{30mm}\cbox{red} TUNA\quad
   \cbox{magenta} CSK\quad
   \cbox{blue} MIL\quad
   \cbox{black} TLD\quad
   \cbox{green} Frag\quad
}%
\label{fig:snap}
\end{figure}

To better demonstrate the performance of TUNA, snapshots for some challenging video sequences are presented in Figure \ref{fig:snap}. Note that TUNA tracks successfully object in long video sequences like \textit{doll} and \textit{lemming} that contain more than $1000$ frames. This is because of the property of the anchor point model that remembers the holistic appearance of the object. Moreover, the keypoints that are matched frequently with higher LT and ST consistencies, help to track the object till the end of the sequence. Moreover, the parameter adaptation of LT consistency associated with keypoints in the model, helps to retain relevant features and remove unreliable features from the model. TUNA results can be found at \href{https://sites.google.com/view/tunatc/tuna}{https://sites.google.com/view/tunatc/tuna}.

\section{Conclusion}\label{sec:con}

In this paper, an online adaptive model-free tracker with a novel anchor point appearance model is proposed. The keypoints are assembled into anchor point features that are arranged in a star graph-like structure with the object center. All the anchor points in the structure votes for the object center and the object localization is done by analyzing the maximum of these voting scores by every keypoint in a score matrix. Our results prove that the anchor point model with constrained structure, acts as robust feature for visual object tracking specifically for tracking objects in low resolution, motion blur, or having deformation or abrupt motion. The voting by a keypoint with a Gaussian helps to tackle the deformation of the object. The dynamic adaptation of long-term consistency and short-term consistency of a keypoint helps in stable and accurate object localization. For the adaptation of scale, a new keypoint pairwise distance measure is proposed. It does not involve complex geometrical or rotation calculation unlike existing methods. Finally, the crucial update of the system is governed by finding similarity of the local pixel level binary features and global weighted color features reference models. Along with this, the features are added and removed from the anchor point appearance model based on their LT consistencies. Nevertheless, the robustness of the proposed tracking approach relies on the keypoint detection step. An interesting direction for future work is to extend the proposed tracker with a detection framework, which may improve further the performance of the tracker.

\begin{acknowledgements}
This work was supported in part by FRQ-NT team grant \#167442 and by REPARTI (Regroupement pour l'\'etude des environnements partag\'es intelligents r\'epartis) FRQ-NT strategic cluster.
\end{acknowledgements}

\begin{table}[t]
\caption{Comparison with state-of-the-art trackers on videos having attributes: Motion Blur (MB), Fast Motion (FM), Background Clutter (BC), Deformation (DEF), Illumination Variation (IV), In-plane Rotation (IPR), Low Resolution (LR), Occlusion (OCC), Out-of-plane-Rotation (OPR), Out-of-View (OV), Scale Variation (SV). The bold italic represents the best results and bold represents the second best results.  }
%\begin{tabular*}{\textwidth}{l@{\extracolsep\fill}ll|ll|ll|ll|ll|ll|ll}
  \begin{tabular*}{\textwidth}{p{1.78cm}p{0.5cm}p{0.5cm}p{0.5cm}p{0.5cm}p{0.5cm}p{0.5cm}p{0.5cm}p{0.5cm}p{0.5cm}p{0.5cm}p{0.5cm}}
    \toprule

   &   && & Overall && Precision \\
 Video Att.   & MB & FM & BC & DEF & IV & IPR & LR & OCC & OPR & OV & SV \\ 
      \midrule
    TUNA\textbf{(Prop.)}& \textbf{0.476} & \textbf{0.452} & 0.348  & \textbf{0.487} & 0.415 & 0.463& \textit{\textbf{0.438}}& 0.486&0.487& \textbf{0.474}& 0.512\\
    MIL \cite{MIL} & 0.338 & 0.382 & \textbf{0.450} & 0.447 & 0.387 & 0.448& 0.168& 0.427&0.461&0.390&0.462\\
    CPF \cite{CPF}& 0.298 & 0.365 & 0.402 & \textit{\textbf{0.488}} & 0.386 & 0.456 & 0.134 & \textbf{0.501}& \textbf{0.510} &0.455&0.464\\
    CSK \cite{CSK}& 0.346& 0.362 &\textit{\textbf{0.534}} & 0.440 & 0.469& 0.513 & \textit{0.437} &0.475&0.506&0.361&0.494\\
    KMS \cite{KMS}& 0.372& 0.359 & 0.391 & 0.404 & 0.384& 0.375 & 0.232&0.401&0.401&0.385&0.408\\
    SemiT \cite{SemiT}& 0.339 & 0.352 & 0.368 & 0.421 & 0.309 & 0.371 & 0.432& 0.391&0.383&0.314&0.376\\
    CT \cite{CT}& 0.316 & 0.333 & 0.327 & 0.418 & 0.352 & 0.368& 0.138& 0.406&0.390&0.348&0.419\\
    BSBT \cite{BSBT}& 0.330 & 0.329 & 0.329 & 0.372 &0.324& 0.388 & 0.244& 0.393&0.400&0.415&0.345\\
    Frag \cite{Adam06robustfragments-based}& 0.274& 0.323 & 0.404 & 0.444 & 0.320& 0.388 & 0.147& 0.441&0.426&0.324&0.379\\
    SMS \cite{Mean-Shift} & 0.299& 0.321 & 0.327 & 0.417 & 0.346 & 0.332 & 0.170 & 0.402&0.401&0.337&0.400\\
    TLD \cite{MIL}& \textit{\textbf{0.482}}& \textit{\textbf{0.517}} & 0.420 & 0.469& \textbf{0.497} & \textit{\textbf{0.545}} & 0.339& \textit{\textbf{0.518}} & \textit{\textbf{0.546}}& \textit{\textbf{0.553}}& \textit{\textbf{0.562}}\\
    ORIA \cite{ORIA} & 0.246 & 0.276 & 0.377 & 0.342& 0.408 & 0.479& 0.236&0.431&0.466&0.323&0.431\\
    ASLA \cite{ASLA}& 0.283& 0.270 & 0.484 &  0.426& \textit{\textbf{0.499}}& \textbf{0.501}& 0.174&0.444&0.500&0.322& \textit{\textbf{0.539}}\\
    IVT \cite{IVT}& 0.220 & 0.219 & 0.395 &0.389& 0.387 &0.435 & 0.272&0.430&0.435&0.290&0.473\\
    \bottomrule
  \end{tabular*}
\label{tab:CLE}
\end{table}

\bibliographystyle{spbasic}      % basic style, author-year citations
\bibliography{egbib}   % name your BibTeX data base

\begin{thebibliography}{42}
\providecommand{\natexlab}[1]{#1}
\providecommand{\url}[1]{{#1}}
\providecommand{\urlprefix}{URL }
\expandafter\ifx\csname urlstyle\endcsname\relax
  \providecommand{\doi}[1]{DOI~\discretionary{}{}{}#1}\else
  \providecommand{\doi}{DOI~\discretionary{}{}{}\begingroup
  \urlstyle{rm}\Url}\fi
\providecommand{\eprint}[2][]{\url{#2}}

\bibitem[{Adam et~al(2006)Adam, Rivlin, and
  Shimshoni}]{Adam06robustfragments-based}
Adam A, Rivlin E, Shimshoni I (2006) Robust fragments-based tracking using the
  integral histogram. In: In IEEE Conf. Computer Vision and Pattern Recognition
  (CVPR), pp 798--805

\bibitem[{Babenko et~al(2011)Babenko, Yang, and Belongie}]{MIL}
Babenko B, Yang MH, Belongie S (2011) {Robust Object Tracking with Online
  Multiple Instance Learning}. IEEE Transactions on Pattern Analysis and
  Machine Intelligence 33(8):1619--1632

\bibitem[{Bouachir and Bilodeau(2014)}]{6836011}
Bouachir W, Bilodeau GA (2014) Structure-aware keypoint tracking for partial
  occlusion handling. In: IEEE Winter Conference on Applications of Computer
  Vision (WACV), 2014, pp 877--884

\bibitem[{Cai et~al(2013)Cai, Wen, Yang, Lei, and Li}]{DBLP:conf/accv/2012-3}
Cai Z, Wen L, Yang J, Lei Z, Li S (2013) Structured visual tracking with
  dynamic graph. In: Computer Vision – ACCV 2012, Lecture Notes in Computer
  Science, vol 7726, Springer Berlin Heidelberg, pp 86--97

\bibitem[{Chakravorty et~al(2015)Chakravorty, Bilodeau, and Granger}]{ctse}
Chakravorty T, Bilodeau GA, Granger E (2015) Contextual object tracker with
  structure encoding. In: 2015 IEEE International Conference on Image
  Processing (ICIP), pp 4937--4941

\bibitem[{Collins(2003)}]{Mean-Shift}
Collins R (2003) Mean-shift blob tracking through scale space. In: IEEE
  Computer Society Conference on Computer Vision and Pattern Recognition, 2003.
  Proceedings., vol~2, pp II--234--40 vol.2

\bibitem[{Comaniciu et~al(2000)Comaniciu, Ramesh, and Meer}]{854761}
Comaniciu D, Ramesh V, Meer P (2000) Real-time tracking of non-rigid objects
  using mean shift. In: Proceedings. IEEE Conference on Computer Vision and
  Pattern Recognition, 2000., vol~2, pp 142--149 vol.2

\bibitem[{Comaniciu et~al(2003)Comaniciu, Ramesh, and Meer}]{KMS}
Comaniciu D, Ramesh V, Meer P (2003) Kernel-based object tracking. IEEE
  Transactions on Pattern Analysis and Machine Intelligence 25(5):564--577

\bibitem[{Danelljan et~al(2014)Danelljan, Khan, Felsberg, and van~de
  Weijer}]{Danelljan_2014_CVPR}
Danelljan M, Khan F, Felsberg M, van~de Weijer J (2014) Adaptive color
  attributes for real-time visual tracking. In: IEEE Conference on Computer
  Vision and Pattern Recognition (CVPR), 2014, pp 1090--1097

\bibitem[{Felzenszwalb et~al(2010)Felzenszwalb, Girshick, McAllester, and
  Ramanan}]{Felzenszwalb:2010:ODD:1850486.1850574}
Felzenszwalb PF, Girshick RB, McAllester D, Ramanan D (2010) Object detection
  with discriminatively trained part-based models. IEEE Trans Pattern Anal Mach
  Intell 32(9):1627--1645

\bibitem[{Grabner et~al(2006)Grabner, Grabner, and
  Bischof}]{BMVC.20.6:abbreviated}
Grabner H, Grabner M, Bischof H (2006) Real-time tracking via on-line boosting.
  In: Proc. BMVC, pp 6.1--6.10

\bibitem[{Grabner et~al(2008)Grabner, Leistner, and Bischof}]{SemiT}
Grabner H, Leistner C, Bischof H (2008) Semi-supervised on-line boosting for
  robust tracking. In: Proceedings of the 10th European Conference on Computer
  Vision: Part I, Springer-Verlag, Berlin, Heidelberg, ECCV '08, pp 234--247

\bibitem[{Hare et~al(2011)Hare, Saffari, and Torr}]{conf/iccv/HareST11}
Hare S, Saffari A, Torr P (2011) Struck: Structured output tracking with
  kernels. In: IEEE International Conference on Computer Vision (ICCV), 2011,
  pp 263--270

\bibitem[{Henriques et~al(2012)Henriques, Caseiro, Martins, and Batista}]{CSK}
Henriques JaF, Caseiro R, Martins P, Batista J (2012) Exploiting the circulant
  structure of tracking-by-detection with kernels. In: Proceedings of the 12th
  European Conference on Computer Vision - Volume Part IV, Springer-Verlag,
  Berlin, Heidelberg, ECCV'12, pp 702--715

\bibitem[{Henriques et~al(2015)Henriques, Caseiro, Martins, and
  Batista}]{henriques2015tracking}
Henriques JF, Caseiro R, Martins P, Batista J (2015) High-speed tracking with
  kernelized correlation filters. IEEE Transactions on Pattern Analysis and
  Machine Intelligence

\bibitem[{Jia et~al(2012)Jia, Lu, and Yang}]{ASLA}
Jia X, Lu H, Yang MH (2012) Visual tracking via adaptive structural local
  sparse appearance model. In: IEEE Conference on Computer Vision and Pattern
  Recognition (CVPR), 2012, pp 1822--1829

\bibitem[{Kalal et~al(2012)Kalal, Mikolajczyk, and Matas}]{TLD}
Kalal Z, Mikolajczyk K, Matas J (2012) Tracking-learning-detection. IEEE
  Transactions on Pattern Analysis and Machine Intelligence 34(7):1409--1422

\bibitem[{Kwon and Lee(2011)}]{6126369}
Kwon J, Lee KM (2011) Tracking by sampling trackers. In: IEEE International
  Conference on Computer Vision (ICCV), 2011, pp 1195--1202

\bibitem[{Leutenegger et~al(2011)Leutenegger, Chli, and
  Siegwart}]{Leutenegger:2011:BBR:2355573.2356277}
Leutenegger S, Chli M, Siegwart RY (2011) Brisk: Binary robust invariant
  scalable keypoints. In: Proceedings of the 2011 International Conference on
  Computer Vision, IEEE Computer Society, Washington, DC, USA, ICCV '11, pp
  2548--2555

\bibitem[{Liu et~al(2011)Liu, Huang, Yang, and
  Kulikowsk}]{BaiyangLiu:2011:RTU:2191740.2191956}
Liu B, Huang J, Yang L, Kulikowsk C (2011) Robust tracking using local sparse
  appearance model and k-selection. In: Proceedings of the 2011 IEEE Conference
  on Computer Vision and Pattern Recognition, IEEE Computer Society,
  Washington, DC, USA, CVPR '11, pp 1313--1320

\bibitem[{Lowe(2004)}]{sift}
Lowe DG (2004) Distinctive image features from scale-invariant keypoints. Int J
  Comput Vision 60(2):91--110

\bibitem[{Matthews et~al(2004)Matthews, Ishikawa, and Baker}]{1288530}
Matthews I, Ishikawa T, Baker S (2004) The template update problem. IEEE
  Transactions on Pattern Analysis and Machine Intelligence 26(6):810--815

\bibitem[{Mei and Ling(2009)}]{DBLP:conf/iccv/MeiL09}
Mei X, Ling H (2009) Robust visual tracking using l1 minimization. In: {IEEE}
  12th International Conference on Computer Vision, {ICCV} 2009, pp 1436--1443

\bibitem[{Nebehay and Pflugfelder(2014)}]{Nebehay2014WACV}
Nebehay G, Pflugfelder R (2014) Consensus-based matching and tracking of
  keypoints for object tracking. In: IEEE Winter Conference on Applications of
  Computer Vision, 2014, IEEE

\bibitem[{Ortiz(2012)}]{Ortiz:2012:FFR:2354409.2354903}
Ortiz R (2012) Freak: Fast retina keypoint. In: Proceedings of the 2012 IEEE
  Conference on Computer Vision and Pattern Recognition (CVPR), IEEE Computer
  Society, Washington, DC, USA, CVPR '12, pp 510--517

\bibitem[{P{\'e}rez et~al(2002)P{\'e}rez, Hue, Vermaak, and Gangnet}]{CPF}
P{\'e}rez P, Hue C, Vermaak J, Gangnet M (2002) Color-based probabilistic
  tracking. In: Proceedings of the 7th European Conference on Computer
  Vision-Part I, Springer-Verlag, London, UK, UK, ECCV '02, pp 661--675

\bibitem[{Possegger et~al(2015)Possegger, Mauthner, and Bischof}]{7298823}
Possegger H, Mauthner T, Bischof H (2015) In defense of color-based model-free
  tracking. In: Computer Vision and Pattern Recognition (CVPR), 2015 IEEE
  Conference on, pp 2113--2120

\bibitem[{Ross et~al(2008)Ross, Lim, Lin, and Yang}]{IVT}
Ross DA, Lim J, Lin RS, Yang MH (2008) Incremental learning for robust visual
  tracking. Int Journal of Computer Vision 77(1-3):125--141

\bibitem[{Shi and Tomasi(1993)}]{Shi:1993:GFT:866676}
Shi J, Tomasi C (1993) Good features to track. Tech. rep., Ithaca, NY, USA

\bibitem[{Smeulders et~al(2014)Smeulders, Chu, Cucchiara, Calderara, Dehghan,
  and Shah}]{6671560}
Smeulders A, Chu D, Cucchiara R, Calderara S, Dehghan A, Shah M (2014) Visual
  tracking: An experimental survey. IEEE Transactions on Pattern Analysis and
  Machine Intelligence 36(7):1442--1468

\bibitem[{St-Charles and Bilodeau(2014)}]{6836059}
St-Charles PL, Bilodeau GA (2014) Improving background subtraction using local
  binary similarity patterns. In: IEEE Winter Conference on Applications of
  Computer Vision (WACV), 2014, pp 509--515

\bibitem[{Stalder et~al(2009)Stalder, Grabner, and v.~Gool}]{BSBT}
Stalder S, Grabner H, v~Gool L (2009) Beyond semi-supervised tracking: Tracking
  should be as simple as detection, but not simpler than recognition. In: IEEE
  12th International Conference on Computer Vision Workshops (ICCV Workshops),
  2009, pp 1409--1416

\bibitem[{Wang and Yeung(2013)}]{NIPS2013_5192}
Wang N, Yeung DY (2013) Learning a deep compact image representation for visual
  tracking. In: Advances in Neural Information Processing Systems 26, pp
  809--817

\bibitem[{Wang et~al(2015)Wang, Li, Gupta, and
  Yeung}]{DBLP:journals/corr/WangLGY15}
Wang N, Li S, Gupta A, Yeung D (2015) Transferring rich feature hierarchies for
  robust visual tracking. CoRR abs/1501.04587

\bibitem[{Wang et~al(2011)Wang, Lu, Yang, and
  Yang}]{ShuWang:2011:ST:2355573.2356430}
Wang S, Lu H, Yang F, Yang MH (2011) Superpixel tracking. In: Proceedings of
  the 2011 International Conference on Computer Vision, IEEE Computer Society,
  Washington, DC, USA, ICCV '11, pp 1323--1330

\bibitem[{Wu et~al(2012)Wu, Shen, and Ling}]{ORIA}
Wu Y, Shen B, Ling H (2012) Online robust image alignment via iterative convex
  optimization. In: CVPR, IEEE Computer Society, pp 1808--1814

\bibitem[{Wu et~al(2013)Wu, Lim, and Yang}]{Wu_2013_CVPR}
Wu Y, Lim J, Yang MH (2013) Online object tracking: A benchmark. In: The IEEE
  Conference on Computer Vision and Pattern Recognition (CVPR)

\bibitem[{Yang et~al(2009)Yang, Wu, and Hua}]{4538230}
Yang M, Wu Y, Hua G (2009) Context-aware visual tracking. IEEE Transactions on
  Pattern Analysis and Machine Intelligence 31(7):1195--1209

\bibitem[{Yilmaz et~al(2006)Yilmaz, Javed, and
  Shah}]{Yilmaz:2006:OTS:1177352.1177355}
Yilmaz A, Javed O, Shah M (2006) Object tracking: A survey. ACM Comput Surv
  38(4)

\bibitem[{Yoon et~al(2012)Yoon, Kim, and Yoon}]{conf/eccv/YoonKY12}
Yoon JH, Kim DY, Yoon KJ (2012) Visual tracking via adaptive tracker selection
  with multiple features. In: ECCV (4), Springer, Lecture Notes in Computer
  Science, vol 7575, pp 28--41

\bibitem[{Zhang et~al(2012)Zhang, Zhang, and Yang}]{CT}
Zhang K, Zhang L, Yang MH (2012) Real-time compressive tracking. In:
  Proceedings of the 12th European Conference on Computer Vision - Volume Part
  III, Springer-Verlag, Berlin, Heidelberg, ECCV'12, pp 864--877

\bibitem[{Zhong et~al(2012)Zhong, Lu, and Yang}]{SCMT}
Zhong W, Lu H, Yang MH (2012) Robust object tracking via sparsity-based
  collaborative model. In: IEEE Conference on Computer Vision and Pattern
  Recognition (CVPR), 2012, pp 1838--1845

\end{thebibliography}

%\begin{thebibliography}{}
% etc
%\end{thebibliography}

\section{Author Biographies}
%\leavevmode

%\vbox{%
%\begin{wrapfigure}{l}{80pt}
%{[{\includegraphics[width=1in,height=1.25in,clip,keepaspectratio]{Tanushri_Chakravorty.jpg}}]\vspace*{100pt}}%
%\end{wrapfigure}
%\noindent\small 
%{\bf Tanushri Chakravorty} received her M.Tech degree in Bio-Medical Instrumentation Eng. at College of Engineering Pune, University of Pune, India in 2011. She is currently pursuing PhD. in Computer Eng. at Polytechnique Montreal. Her research interests lie in computer vision with focus on object tracking and video surveillance.\vadjust{\vspace{40pt}}}

\vspace{20pt}
\noindent\small 
{\bf Tanushri Chakravorty} received her M.Tech degree in Bio-Medical Instrumentation Eng. at College of Engineering Pune, University of Pune, India in 2011. She is currently pursuing PhD. in Computer Eng. at Polytechnique Montreal. Her research interests lie in computer vision with focus on object tracking and video surveillance.
\vspace{20pt}

\noindent\small 
{\bf Guillaume-Alexandre Bilodeau}
(M'10) received the B.Sc.A. degree in computer engineering and the Ph.D.\ degree in electrical engineering from Universit\'e Laval, Canada, in 1997 and 2004, respectively. In 2004, he was appointed Assistant Professor at Polytechnique Montr\'eal, Canada, where he is now Full professor since 2014. His research interests encompass image and video processing, video surveillance, object tracking, segmentation, and medical applications of computer vision. Dr. Bilodeau is a member of the REPARTI research network.
\vspace{20pt}

\noindent\small 
{\bf \'{E}ric Granger}
earned Ph.D. in EE from Polytechnique Montr\'{e}al in 2001, and worked as a Defense Scientist at DRDC-Ottawa (1999-2001), and in R\& D with Mitel Networks (2001-04). He joined the \'{E}cole de technologie sup\'{e} rieure (Universit\'{e} du Qu\'{e} bec), Montreal, in 2004, where he is presently Professor and Director of LIVIA, a research laboratory on computer vision and artificial intelligence. His research focuses on adaptive pattern recognition, machine learning, computer vision and computational intelligence.

\end{document}